\documentclass[10pt,twocolumn,letterpaper]{article}

\usepackage{cvpr}              

\usepackage[dvipsnames]{xcolor}

\definecolor{cvprblue}{rgb}{0.21,0.49,0.74}
\usepackage[pagebackref,breaklinks,colorlinks,citecolor=cvprblue]{hyperref}

\usepackage{xcolor}
\usepackage{multirow, booktabs, colortbl, graphicx}
\definecolor{lightgray}{gray}{.9}
\usepackage{algorithm, algorithmic}
\usepackage[accsupp]{axessibility}

\def\paperID{3301} 
\def\confName{CVPR}
\def\confYear{2024}

\title{Cross-Domain Few-Shot Segmentation\\via Iterative Support-Query Correspondence Mining}

\author{Jiahao Nie$^{1,2}$\thanks{Equal contribution} \quad
Yun Xing$^2$\footnotemark[1] \quad
Gongjie Zhang$^3$ \quad
Pei Yan$^{4,2}$ \quad
Aoran Xiao$^2$ \\
Yap-Peng Tan$^2$ \quad
Alex C. Kot$^2$ \quad
Shijian Lu$^2$\thanks{Corresponding author}\\
$^1$Interdisciplinary Graduate Programme, Nanyang Technological University \\
$^2$Nanyang Technological University ~\quad
$^3$Black Sesame Technologies \\
$^4$Huazhong University of Science and Technology\\
{\tt\small \{jiahao007,xing0047\}@e.ntu.edu.sg \ gjz@ieee.org \ yanpei@hust.edu.cn} \\
{\tt\small\{aoran.xiao,eyptan,eackot,shijian.lu\}@ntu.edu.sg}
}

\begin{document}
\maketitle
\begin{abstract}

Cross-Domain Few-Shot Segmentation (CD-FSS) poses the challenge of segmenting novel categories from a distinct domain using only limited exemplars. In this paper, we undertake a comprehensive study of CD-FSS and uncover two crucial insights: (i) the necessity of a fine-tuning stage to effectively transfer the learned meta-knowledge across domains, and (ii) the overfitting risk during the naïve fine-tuning due to the scarcity of novel category examples. With these insights, we propose a novel cross-domain fine-tuning strategy that addresses the challenging CD-FSS tasks. We first design Bi-directional Few-shot Prediction (BFP), which establishes support-query correspondence in a bi-directional manner, crafting augmented supervision to reduce the overfitting risk. Then we further extend BFP into Iterative Few-shot Adaptor (IFA), which is a recursive framework to capture the support-query correspondence iteratively, targeting maximal exploitation of supervisory signals from the sparse novel category samples. Extensive empirical evaluations show that our method significantly outperforms the state-of-the-arts (+7.8\%), which verifies that IFA tackles the cross-domain challenges and mitigates the overfitting simultaneously. The code is available at: \href{https://github.com/niejiahao1998/IFA}{https://github.com/niejiahao1998/IFA}.
\end{abstract}    
\section{Introduction}
\label{sec:intro}

Few-Shot Segmentation (FSS) aims to segment novel categories based on very limited support exemplars, typically by transferring category-agnostic knowledge learned from abundant base categories to novel categories~\cite{snell2017prototypical,dong2018few,tian2020prior,fan2022self,min2021hypercorrelation,peng2023hierarchical}. Similar to other few-shot tasks~\cite{chen2019closer,hou2019cross,kang2021relational,kang2019few,zhang2021pnpdet,xiao2022few,zhang2022meta}, FSS is generally addressed with meta-learning~\cite{chen2019closer,tseng2019cross,boudiaf2021few,lu2021simpler,wang2022remember} and learns generalizable category correspondence from constructed support-query pairs.

\begin{figure}[t]
    \centering
    \includegraphics[width=1.0\linewidth]{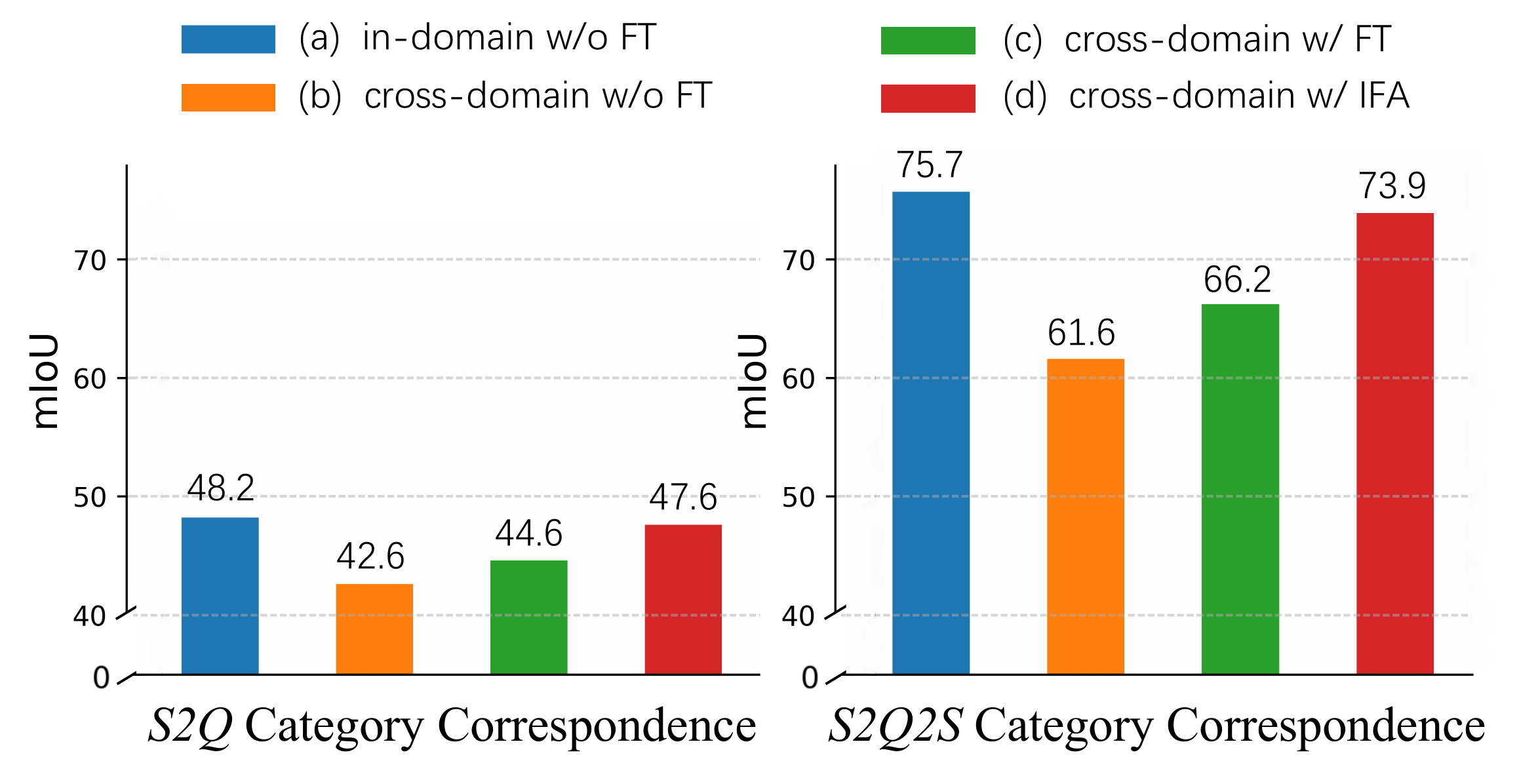}
    \vspace{-6.5mm}
    \caption{We investigate two types of category correspondence, \textbf{left:} \textit{Support-to-Query (S2Q)} and \textbf{right:} \textit{Support-to-Query-to-Support (S2Q2S)} under four experimental setups (a-d). (a) In-domain performances without fine-tuning (FT) set oracle baselines for Cross-Domain Few-Shot Segmentation (CD-FSS). (b) Cross-domain results without fine-tuning suffer from severe performance drops, which verifies the necessity of bridging domain gap for CD-FSS. (c) Cross-domain setups with naïve fine-tuning only bring small performance gains, which is attributed to the overfitting risk of CD-FSS fine-tuning. Notably, there also underlies rich unexplored category correspondence in \textit{S2Q2S}. (d) Cross-domain setup with our proposed Iterative Few-Shot Adaptor (IFA) achieve significant performance gains. IFA comprehensively exploits maximum information content in the given data by capturing both \textit{S2Q} and \textit{S2Q2S} category correspondence during fine-tuning.}
    \label{fig:motivation}
    \vspace{-4.5mm}
\end{figure}

Current state-of-the-art FSS approaches~\cite{fan2022self,peng2023hierarchical} have demonstrated impressive performance when novel categories fall in the same domain with base categories, yet still suffer from severe performance drop when a clear domain gap is present~\cite{lei2022cross} (\eg, base categories from Pascal VOC~\cite{everingham2010pascal}, and novel categories from Deepglobe~\cite{demir2018deepglobe}). To bridge this gap, the task of Cross-Domain Few-Shot Segmentation (CD-FSS) has been explored for generalizing the ability of FSS beyond its own domain~\cite{lei2022cross,wang2022remember}. Nevertheless, directly applying a meta-learned model to target domain is incapable of mitigating this challenge~\cite{lei2022cross}, largely because the category correspondence learned under meta-learning still biases toward the source domain.

\begin{figure}[t]
  \centering
    \includegraphics[width=0.85\linewidth]{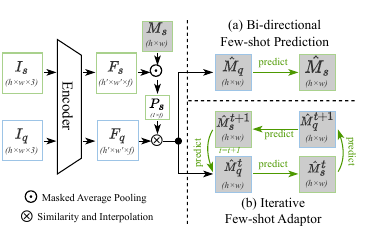}
    \vspace{-7mm}
    \caption{Illustration of our designs for CD-FSS: (a) Bi-directional Few-shot Prediction, and (b) Iterative Few-shot Adaptor. $I_s$ and $I_q$ denote support and query images respectively. $F_s$ and $F_q$ denote the corresponding support and query features as extracted by the \textit{Encoder}. $M_s$ denotes the support mask, and $P_s$ denotes the generated support prototype.}
    \label{fig:iteration}
    \vspace{-5mm}
\end{figure}

We conduct a series of in-domain and cross-domain FSS experiments to verify our hypothesis and explore solutions\footnote{Experiments in Fig.~\ref{fig:motivation} are conducted by meta-learning on the same amount of base data and generalizing to the same novel categories (in-domain: Deepglobe, cross-domain: Pascal VOC to Deepglobe).}. Specifically, we investigate two types of category correspondence, \textit{Support-to-Query (S2Q)} and \textit{Support-to-Query-to-Support (S2Q2S)}, to measure the generalization capability under different setups. Concretely, \textit{S2Q} uses support exemplars to segment query images, while \textit{S2Q2S} uses \textit{S2Q} outcomes to predict back on support images. We introduce category correspondence of \textit{S2Q2S} because it is not directly optimized in meta-learning and has a lower risk of overfitting. As shown in Fig.~\ref{fig:motivation}(a), in-domain performances without fine-tuning set oracle performances for cross-domain setups. Notably, cross-domain performances (Fig.~\ref{fig:motivation}(b)) are lower than the corresponding in-domain performances (Fig.~\ref{fig:motivation}(a)), verifying our hypothesis that the learned category correspondence has a clear bias toward the source domain. To generalize the category correspondence beyond its own domain, it is straightforward to introduce a fine-tuning stage~\cite{guo2020broader,lei2022cross,wang2022revisit}. However, naïve fine-tuning on novel category exemplars (Fig.~\ref{fig:motivation}(c)) only brings small performance gains. We attribute this phenomenon to overfitting caused by the limited accessible samples from the target domain, which is a core challenge in CD-FSS.

To address the overfitting challenge, we propose a novel fine-tuning strategy for CD-FSS. We first design Bi-directional Few-shot Prediction (BFP), which captures both \textit{S2Q} and \textit{S2Q2S} category correspondence simultaneously during meta-learning (Fig.~\ref{fig:iteration}(a)). This design leverages support masks as additional supervision to mitigate the overfitting risk. Building upon BFP, we further design Iterative Few-shot Adaptor (IFA), as illustrated in Fig.~\ref{fig:iteration}(b). Specifically, IFA iteratively conducts BFP and constructs supervision signals for predictions in every iteration. Hence, IFA comprehensively exploits the supervision from few-shot target exemplars, thereby mining extensive support-query correspondence during fine-tuning. Extensive experiments on four CD-FSS benchmarks show the effectiveness of our designs, especially in mitigating the overfitting challenge.

\noindent Our contributions can be summarized in four major aspects:
\begin{itemize}[topsep=1pt, itemsep=1pt, parsep=1pt]
    \item [$\bullet$]We conduct a comprehensive study on the CD-FSS challenge, verifying the necessity of a fine-tuning stage and the overfitting risk during the naïve fine-tuning.
    \item [$\bullet$]We design a Bi-directional Few-shot Prediction (BFP) module to establish support-query correspondence, which leverages extensive supervision signals to mitigate overfitting risk during fine-tuning.
    \item [$\bullet$]We extend BFP to Iterative Few-shot Adaptor (IFA) in a recursive framework, fully exploiting the supervision signals from limited samples with iterative support-query correspondence mining.
    \item [$\bullet$]Our method tackles the cross-domain and overfitting challenges simultaneously and surpasses the state-of-the-art methods by large margins.
\end{itemize}
\section{Related Work}

\noindent\textbf{Few-Shot Segmentation (FSS)}~\cite{shaban2017one,dong2018few,wang2019panet,tian2020prior,zhang2021few,fan2022self,peng2023hierarchical} performs segmentation for novel categories with only a few annotations, which has been studied extensively. Most existing works can be categorized into two types. Prototype-based methods~\cite{snell2017prototypical,dong2018few,wang2019panet,tian2020prior,li2021adaptive,liu2022intermediate,lang2022learning,lang2023base} perform segmentation by similarities between all query features and support prototypes. In contrast to prototype-based methods, affinity-based ones~\cite{lu2021simpler,min2021hypercorrelation,zhang2021few,peng2023hierarchical} mine dense correspondence between query and support features, which rely on rich contextual information. Although the aforementioned methods are well-established, their robustness under cross-domain setups is under-examined. In Cross-Domain Few-Shot Segmentation (CD-FSS) setups, we highlight that: \textit{(i)} existing prototype-based methods yield unsatisfactory performance because learned category correspondence is hardly generalized to target domains; \textit{(ii)} existing affinity-based methods are also unsuitable, as affected by much irrelevant information~\cite{wang2022adaptive} when performing fine-tuning with limited target data. Different from prior attempts, our design demonstrates distinct superiority in CD-FSS, leveraging an effective iterative fine-tuning strategy.

\noindent\textbf{Domain Adaptive Segmentation (DAS)} is a paradigm to mitigate costly annotation and domain gap issues. Existing DA methods mitigate these issues by: \textit{(i)} employing a discriminator to alleviate differences between different domains at output-level~\cite{tsai2018learning,vu2019advent,chen2019domain,guan2021scale} or feature-level~\cite{hoffman2016fcns,hong2018conditional,luo2019significance,zhang2023detr}; \textit{(ii)} re-training a model learned from source domain with pseudo labels derived from target domain predictions~\cite{zou2018unsupervised,luo2021unsupervised,xing2022domain,guan2021domain}; or \textit{(iii)} generating source-style target data and reducing domain gap at input-level~\cite{hoffman2018cycada,xiao2022polarmix}. Recently, FSS in cross-domain setup has been explored~\cite{lu2021simpler,boudiaf2021few,wang2022remember}. Lei~\etal~\cite{lei2022cross} first propose the Cross-Domain Few-Shot Segmentation (CD-FSS) framework to transfer the trained models to different low-resource domains. Then Fan~\etal~\cite{fan2023darnet} and Huang~\etal~\cite{huang2023restnet} design approaches from test-time fine-tuning and knowledge-transfer aspects, respectively. Motivated by these pioneer works, Chen~\etal~\cite{chen2024pixel} propose a universal method to solve in-domain and cross-domain FSS tasks together. CD-FSS is different from DAS in two folds: \textit{(i)} when adapting to the target domain, unlabeled data are abundant in DAS but only limited support data is given in CD-FSS, posing risks to overfitting; \textit{(ii)} DAS assumes source and target label spaces are the same, while the labels spaces of source and target domains in CD-FSS are disjoint. Distinct from the above efforts, we focus on designing an effective fine-tuning strategy for few-shot challenge with limited accessible data.

\section{Problem Formulation}\label{sec:formulation}

Cross-Domain Few-Shot Segmentation (CD-FSS) transfers meta-learned capability of segmenting novel categories to new target domains with only a few annotated support images. Please refer to supplementary materials for more description of CD-FSS. By definition, the model is trained on the source domain \boldmath$\mathcal{D}_{source}$ and is evaluated on the target domain $\mathcal{D}_{target}$. Let \unboldmath$\{\mathcal{X}_s,\mathcal{Y}_s\}$ and $\{\mathcal{X}_t, \mathcal{Y}_t\}$ denote the sets in \boldmath$\mathcal{D}_{source}$ and $\mathcal{D}_{target}$ respectively, where \unboldmath$\mathcal{X}$ denotes the data distribution and $\mathcal{Y}$ denotes the label space. The data distribution of source and target domains are different, and there is no overlap between the source and target label spaces, \ie, $\mathcal{X}_s \ne \mathcal{X}_t, \ \mathcal{Y}_s \cap \mathcal{Y}_t = \emptyset$.

Following the meta-learning in~\cite{lei2022cross}, we adopt the episode training strategy. Specifically, each episode is constructed by a support set $S = {\{(I_s^i, M_s^i)\}}_{i=1}^K$ and a query set $Q = \{(I_q, M_q)\}$ within the same category, where $I_s, I_q$ denote the support and query images, and $M_s, M_q$ denote their masks. The framework can be divided into three steps: \textit{(i)} training the model in \boldmath$\mathcal{D}_{source}$ with both \unboldmath$M_s$ and $M_q$; \textit{(ii)} fine-tuning the trained model to \boldmath$\mathcal{D}_{target}$ with only \unboldmath$M_s$; and \textit{(iii)} testing the adapted model in \boldmath$\mathcal{D}_{target}$.
\begin{figure*}[t]
  \centering 
    \includegraphics[width=0.97\linewidth]{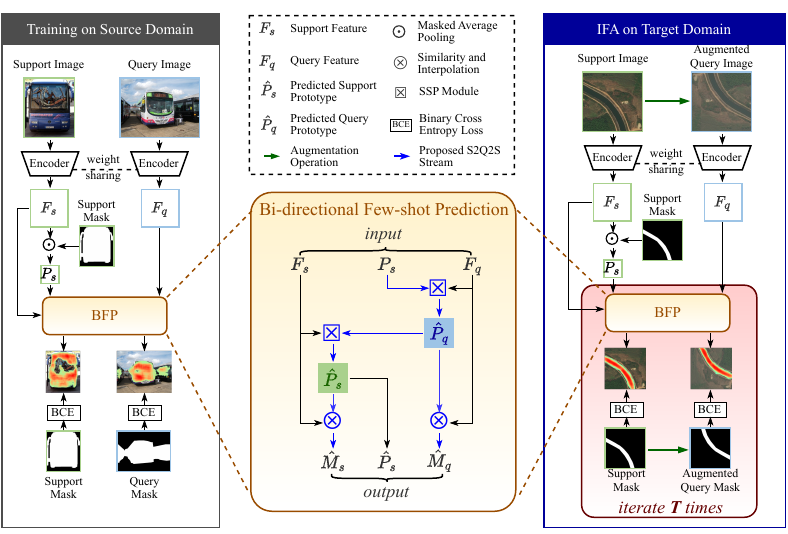}
    \vspace{-3mm}
    \caption{Overall architecture of the proposed Iterative Few-shot Adaptor (IFA), which is composed of two essential steps: training on the source domain, and fine-tuning over the target domain. In the training stage, we only adopt the Bi-directional Few-shot Prediction (BFP) (illustrated in yellow box), which is the fundamental unit of IFA. BFP is composed of both \textit{S2Q} and \textit{S2Q2S} streams together with supervision signals from both sides (blue arrows). In the fine-tuning stage where the target exemplars are extremely scarce, IFA is designed to iterate BFP \textbf{$T$} times, recursively mining the support-query correspondence (illustrated in red box). To show the predictions clearly, we only visualize the region where confidence is higher than 0.5.}
    \label{fig:framework}
    \vspace{-4mm}
\end{figure*}

\section{Method}

To transfer the capability of segmenting novel categories to target domains, we propose a method that mines support-query correspondence iteratively, as illustrated in Fig.~\ref{fig:framework}. The proposed method consists of two major steps: \textit{(i)} training models with Bi-directional Few-shot Prediction (BFP) on the source domain; and \textit{(ii)} fine-tuning trained models to target domains with Iterative Few-shot Adaptator (IFA). The basic pipeline can be formulated as follows: The input support and query images \unboldmath$\{I_{s}, I_{q}\}$ are fed into a weight-shared encoder to extract features $\{F_{s}, F_{q}\}$. Then the support feature $F_s$ and its mask $M_s$ are processed by the masked average pooling to generate support prototype $P_s$.  Finally, using $P_s$, $F_s$, and $F_q$ to predict query masks $\hat{M}_q$. We elaborate our designs as follows: we first revisit SSP method~\cite{fan2022self} in Sec.~\ref{ssec:ssp}, which is used directly in our method. Then, the BFP and IFA are presented in Sec.~\ref{ssec:bfp} and Sec.~\ref{ssec:ifa}, respectively. Finally, we explain how to extend our design into $K$-shot setting in Sec.~\ref{ssec:kshot}.

\subsection{Revisiting of SSP}\label{ssec:ssp}
Motivated by the simple Gestalt principle~\cite{koffka2013principles} that pixels belonging to the same object are more similar than those from different objects, thus the given support may not be a good reference for predicting query mask. The target datasets of Cross-Domain Few-Shot Segmentation (CD-FSS) comply with the Gestalt principle, which is verified in supplementary materials. Fan~\etal propose a Self-Support Prototype (SSP) module to alleviate this problem~\cite{fan2022self}. Firstly, SSP generates support prototype $P_s$ from support feature $F_s$ and mask $M_s$:
\begin{equation}\label{eqn:ps}
    P_{s} = MAP(F_{s}, M_{s}),
\end{equation}
where $MAP$ is masked average pooling operation. Different from traditional prototypical learning~\cite{dong2018few} with direct matching between the support prototype and query image, SSP takes a two-step matching. Concretely, SSP uses a support prototype to find the most similar region in the query image first and then takes such region self-matching in query image to predict the mask. Our method (1-shot) is illustrated in Fig.~\ref{fig:framework}, and we adopt SSP module to predict the query prototype $\hat{P}_q$ from support prototype $P_s$ and query feature $F_q$:
\begin{equation}
    \hat{P}_{q} = SSP(F_{q}, P_{s}).
\end{equation}
Compared with previous methods, SSP yields a more representative prototype for query~\cite{fan2022self}, which is important for our design. More implementation details of applying SSP for our method are shown in supplementary materials.

\subsection{Bi-directional Few-shot Prediction}\label{ssec:bfp}

Considering that only a few support annotations are accessible when fine-tuning to target domains, a well-designed strategy for establishing support-query correspondence is essential. The previous uni-directional \textit{Support-to-Query (S2Q)} prediction is presented below. Predicted support mask $\hat{M}_{s}$ and query mask $\hat{M}_{q}$ can be obtained by:
\begin{equation}\label{eqn:predicted_ms}
    \begin{aligned}
        &\hat{M}_{s} = softmax(Sim(F_{s}, P_{s})),\\
        &\hat{M}_{q} = softmax(Sim(F_{s}, \hat{P}_{q})),
    \end{aligned}
\end{equation}
where $Sim$ is cosine similarity. Then support base loss $\mathcal{L}_{bs}$ and query base loss $\mathcal{L}_{bq}$ are adopted:
\begin{equation}\label{eqn:base_loss}
    \begin{aligned}
        &\mathcal{L}_{bs} = BCE(\hat{M}_{s}, M_{s}),\\
        &\mathcal{L}_{bq} = BCE(\hat{M}_{q}, M_{q}),
    \end{aligned}
\end{equation}
where $BCE$ is the binary cross entropy loss. Because accessible target exemplars are limited, the naïve meta-learning paradigm easily leads to overfitting. Consequently, it is important to introduce extra information via given data.

Therefore, we propose another \textit{Support-to-Query-to-Support (S2Q2S)} stream to conduct more predictions:
\begin{equation}
    \hat{P}_{s'} = SSP(F_{s}, \hat{P}_q),
\end{equation}
where $\hat{P}_{s'}$ is another support prototype predicted from query. Then, we can introduce corresponding loss $\mathcal{L}_{s'}$ via support ground-truth:
\begin{equation}
    \begin{aligned}
        \hat{M}_{s'}& = softmax(Sim(F_{s}, \hat{P}_{s'}),\\
        &\mathcal{L}_{s'} = BCE(\hat{M}_{s'}, M_{s}),\\
    \end{aligned}
\end{equation}
combining $\mathcal{L}_b$ and $\mathcal{L}_{s'}$ helps establishing more robust support-query correspondence.

Our proposed method is supervised by both support and query masks, which are accessible when training models on the source domain. However, query labels are unavailable when fine-tuning to target domains. Consequently, we derive query image $I_{q}$ with corresponding label $M_{q}$ from support image $I_{s}$ and mask $M_{s}$:
\begin{equation}
    I_{q},\ M_{q} = \mathcal{AUG}(I_{s}),\ \mathcal{AUG}(M_{s}),
\end{equation}
where $\mathcal{AUG}$ is an augmentation operation, which is described in Sec.~\ref{ssec:exp_detail}. It is worth noting that we adopt the same transformation operations on $I_{s}$ and $M_{s}$, to ensure that generated $I_{q}$ and $M_{q}$ are matched.

\subsection{Iterative Few-shot Adaptor}\label{ssec:ifa}

Since $I_q$ and $M_q$ are augmented from $I_s$ and $M_s$, the region-of-interest should be same in $I_s$ and $I_q$. If the model is trained well, $\hat{M}_q$ should be accurately predicted from $P_s$, and $\hat{M}_{s'}$ should also be securely predicted from $\hat{P}_q$. Otherwise, iteratively conducting predictions may lead to variant results, which measure the generalization capability of learned category correspondence. To further tackle the challenge brought by data scarcity, conducting BFP iteratively on the given data helps the model learn better support-query correspondence. Assuming we totally have $\textbf{\textit{T}}$ times bi-directional predictions, procedures at the iteration $j$ are:
\begin{equation}\label{eqn:iter}
    \begin{aligned}
        &\hat{P}_{q}^{j+1} = SSP(F_{q}, \hat{P}_{s}^{j}),\\
        &\hat{P}_{s}^{j+1} = SSP(F_{s}, \hat{P}_{q}^{j+1}),\\
        \hat{M}_{q}^{j+1}& = softmax(Sim(F_{s}, \hat{P}_{q}^{j+1})),\\
        \hat{M}_{s}^{j+1}& = softmax(Sim(F_{s}, \hat{P}_{s}^{j+1})).       
    \end{aligned}
\end{equation}
$\hat{P}_{q}$, $\hat{P}_{s'}$, $\hat{M}_{q}$ and $\hat{M}_{s'}$ in Sec.~\ref{ssec:bfp} can be treated as $\hat{P}_{q}^{1}$, $\hat{P}_{s}^{1}$, $\hat{M}_{q}^{1}$ and $\hat{M}_{s}^{1}$ when $j=0$.

In the meantime, prior predicted errors also accumulate and lead to inaccurate outcomes after $\textbf{\textit{T}}$ times iterations. To tackle this problem, we introduce supervision signals $\mathcal{L}_{q}^{j+1}$ and $\mathcal{L}_{s}^{j+1}$ in every iteration:
\begin{equation}\label{eqn:iter_loss}
    \begin{aligned}
        &\mathcal{L}_{q}^{j+1} = BCE(\hat{M}_{q}^{j+1}, M_{q}),\\
        &\mathcal{L}_{s}^{j+1} = BCE(\hat{M}_{s}^{j+1}, M_{s}),
    \end{aligned}
\end{equation}
$\mathcal{L}_{bq}$, $\mathcal{L}_{s'}$ in Sec.~\ref{ssec:bfp} can be treated as $\mathcal{L}_{q}^{1}$ and $\mathcal{L}_{s}^{1}$ when $j=0$, respectively. It is worth noting that we only conduct iterative predictions in the fine-tuning step to mine category correspondence of target domain. Once the model is learned well, we assume that support prototype $\hat{P}_s$ should be representative, thus we only directly adopt support-to-query prediction in the testing stage.

The total loss $\mathcal{L}$ includes all aforementioned losses with different weights:
\begin{equation}\label{eqn:all_loss}
    \begin{aligned}
        \mathcal{L} =& \lambda_{bs}\times\mathcal{L}_{bs}+\lambda_{bq}\times\mathcal{L}_{bq}+\lambda_{s'}\times\mathcal{L}_{s'}\\
        &+\lambda_{i}\times\sum_{j=1}^{\textit{T}-1}(\mathcal{L}_{q}^{j+1}+\mathcal{L}_{s}^{j+1}),
    \end{aligned}
\end{equation}
where $\lambda_{bs}=0.2$, $\lambda_{bq}=1.0$, $\lambda_{s'}=0.4$, and $\lambda_{i}=0.1$. The ablation studies on hyper-parameters are shown in supplementary materials. The overall IFA flow in target domains is presented in Algo.~\ref{algo:ifa}.

\begin{algorithm}[h]
    \caption{Pipeline of our proposed Iterative Few-shot Adaptor (1-shot setup).}\label{algo:ifa}
    \begin{algorithmic}[1]
        \REQUIRE Support image $I_s$, Query image $I_q$, Ground-truth support mask $M_s$, Ground-truth query mask $M_q$, Source-trained model $\mathcal{G}_s$, and Determined iterative times $\textbf{\textit{T}}$ of predictions.
        \ENSURE Adapted target model $\mathcal{G}_t$.
        \STATE $\mathcal{G}_t=\mathcal{G}_s$
        \WHILE {not reach the maximum epoch}
            \STATE \textbf{initialize:}
            \STATE \quad $j=0$, $\mathcal{L}=0$.
            \STATE \textcolor{blue}{/* Derive fundamental information: */}
            \STATE Extract support feature $F_s$ via $\mathcal{G}_t$ and $I_s$.
            \STATE Extract query feature $F_q$ via $\mathcal{G}_t$ and $I_q$.
            \STATE Calculate $P_s$ by Eqn.\ref{eqn:ps}.
            \STATE Predict $\hat{M}_s$ by Eqn.\ref{eqn:predicted_ms}.
            \STATE Calculate $\mathcal{L}_{bs}$ by Eqn.\ref{eqn:base_loss} and add to $\mathcal{L}$.
            \STATE \textcolor{blue}{/* Start iterative predictions: */}
            \STATE $\hat{P}_{s}^{0}=P_s$.
            \WHILE {j\ \textless\ \textbf{\textit{T}}}
                \STATE Calculate $\hat{P}_{q}^{j+1}$ and $\hat{P}_{s}^{j+1}$ by Eqn.\ref{eqn:iter}.
                \STATE Predict $\hat{M}_{q}^{j+1}$ and $\hat{M}_{s}^{j+1}$ by Eqn.\ref{eqn:iter}.
                \STATE Calculate $\mathcal{L}_{q}^{j+1}$ and $\mathcal{L}_{s}^{j+1}$ by Eqn.\ref{eqn:iter_loss} and add to $\mathcal{L}$.
                \STATE $j=j+1$.
            \ENDWHILE
            \STATE \textcolor{blue}{/* Model parameter updating: */}
            \STATE Optimize and update target model $\mathcal{G}_t$.
        \ENDWHILE
    \end{algorithmic}
\end{algorithm}

\subsection{Extension to $K$-shot Setting}\label{ssec:kshot}
In extension to $K$-shot ($K$ \textgreater\ 1) setting, $K$ support images with the corresponding masks $S = {\{(I_s^i, M_s^i)\}}_{i=1}^K$ are given for fine-tuning. We derive $(I_q, M_q)$ from one randomly picked pair $(I_s^i, M_s^i)$. IFA can be quickly and easily extended to the new setting as follows.

As indicated in Sec.~\ref{ssec:bfp}, \textit{S2Q} prediction becomes using averaged $\bar{P}_s = \frac{1}{K}\sum_{i=1}^{K}P_{s}^{i}$ to predict $\hat{P}_q$. Accordingly, we use $\hat{P}_q$ to predict each $\hat{P}_{s'}^{i}$ one by one in the \textit{S2Q2S} prediction procedure. In the iterative design in Sec.~\ref{ssec:ifa}, IFA repeats the above steps $\textbf{\textit{T}}$ times for $K$-shot setting.
\section{Experiment}
\label{sec:exp}

\subsection{Benchmark}
\begin{figure}[ht]
  \centering
    \includegraphics[width=0.88\linewidth]{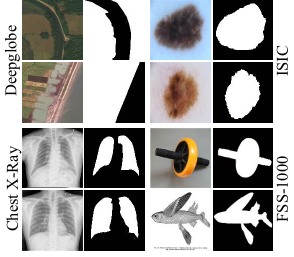}
    \vspace{-2mm}
    \caption{Examples of images and their corresponding ground-truth masks from four target domain datasets, encompassing a diverse range from satellite images and medical screenings to minuscule everyday objects.}
    \label{fig:dataset_sample}
    \vspace{-5mm}
\end{figure}

Following the settings in \cite{lei2022cross}, we utilize PASCAL VOC 2012 dataset~\cite{everingham2010pascal} with SBD augmentation~\cite{hariharan2011semantic} as the source domain for training, which contains 20 daily object categories. Subsequently, we fine-tune the trained model to four target domains, encompassing satellite images, medical screenings, and tiny daily objects. Deepglobe~\cite{demir2018deepglobe} is a satellite image dataset with 7 categories: areas of urban, agriculture, rangeland, forest, water, barren, and unknown. For Cross-Domain Few-Shot Segmentation (CD-FSS), we inherit the same pre-processing in \cite{lei2022cross} to filter out unknown category and cut images into small pieces. ISIC2018~\cite{codella2019skin, tschandl2018ham10000} is an RGB medical image dataset of skin lesions, containing 3 different types of lesions. To augment the diversity of the target domains, we also incorporate a black-white Chest X-Ray medical screening dataset~\cite{candemir2013lung, jaeger2013automatic}, which is collected for Tuberculosis. FSS-1000~\cite{li2020fss} is an everyday object dataset. However, it presents considerable challenges because it contains scarce and tiny objects not present in the source domain. Sample images and their corresponding masks from four target datasets are shown in Fig.~\ref{fig:dataset_sample}.

\subsection{Implementation Details}
\label{ssec:exp_detail}
We adopt the popular ResNet~\cite{he2016deep} pre-trained on ImageNet~\cite{deng2009imagenet} as the backbone. Following baseline SSP~\cite{fan2022self}, we discard the last backbone stage and the last ReLU for better generalization. The model is implemented in PyTorch~\cite{paszke2017automatic}. During training on the source domain, our model is optimized with 0.9 momentum and an initial learning rate of 1e-3. Consistent with PATNet~\cite{lei2022cross}, we resize both support and query images to $400\times400$, for reducing the memory consumption and speeding up the learning process. In the fine-tuning step, the learning rate is set as 5e-4 for Deepglobe, ISIC, and FSS-1000, and 1e-5 for Chest X-Ray. For each dataset, a total of 40 epochs are optimized, with 20 epochs dedicated to training and the remaining for fine-tuning. We randomly adopt a set of transforms function in PyTorch~\cite{paszke2017automatic}, including horizontal-flip, vertical-flip, rotation by 90 degrees, brightness-variation, and hue-variation, for the augmentation operation in Sec.~\ref{ssec:bfp}.

\begin{table*}[ht]
    \renewcommand\arraystretch{1.2}
    \centering
    \begin{tabular}{l|c|cc|cc|cc|cc|cc}
        \toprule[1pt]
        \multicolumn{12}{c}{Source Domain: Pascal VOC 2012 $\to$ Target Domain: Below}\\\hline
        \multirow{2}*{Methods}& \multirow{2}*{Backbone}& \multicolumn{2}{c|}{Deepglobe}& \multicolumn{2}{c|}{ISIC}& \multicolumn{2}{c|}{Chest X-Ray}& \multicolumn{2}{c|}{FSS-1000}& \multicolumn{2}{c}{\textbf{Average}}\\\cline{3-12}
        ~& ~& 1-shot& 5-shot& 1-shot& 5-shot& 1-shot& 5-shot& 1-shot& 5-shot& 1-shot& 5-shot\\\hline
        AMP~\cite{siam2019amp}& \multirow{2}*{Vgg-16}&  37.6& 40.6& 28.4& 30.4& 51.2& 53.0& 57.2& 59.2& 43.6& 45.8\\
        PATNet~\cite{lei2022cross}& ~& 28.7& 34.8& 33.1& 45.8& 57.8& 60.6& 71.6& 76.2& 47.8& 54.4\\\hline
        PGNet~\cite{zhang2019pyramid}& \multirow{10}*{Res-50}& 10.7& 12.4& 21.9& 21.3& 34.0& 28.0& 62.4& 62.7& 32.2& 31.1\\
        PANet~\cite{wang2019panet}& ~& 36.6& 45.3& 25.3& 34.0& 57.8& 69.3& 69.2& 71.7& 47.2 & 55.1\\
        CaNet~\cite{zhang2019canet}& ~& 22.3& 23.1& 25.2& 28.2& 28.4& 28.6& 70.7& 72.0& 36.6& 38.0\\
        RPMMs~\cite{yang2020prototype}& ~& 13.0& 13.5& 18.0& 20.0& 30.1& 30.8& 65.1& 67.1& 31.6& 32.9\\
        PFENet~\cite{tian2020prior}& ~& 16.9& 18.0& 23.5& 23.8& 27.2& 27.6& 70.9& 70.5& 34.6& 35.0\\
        RePRI~\cite{boudiaf2021few}& ~& 25.0& 27.4& 23.3& 26.2& 65.1& 65.5& 71.0& 74.2& 46.1& 48.3\\
        HSNet~\cite{min2021hypercorrelation}& ~& 29.7& 35.1& 31.2& 35.1& 51.9& 54.4& 77.5& 81.0& 47.6& 51.4\\
        PATNet~\cite{lei2022cross}& ~& 37.9& 43.0& 41.2& 53.6& 66.6& 70.2& \underline{78.6}& \underline{81.2}& 56.1& 62.0\\
        SSP$^{*}$~\cite{fan2022self}& ~& \underline{41.3}& \underline{54.2}& \underline{48.6}& \underline{65.4}& \underline{72.6}& \underline{73.0}& 77.0& 79.4& \underline{60.0}& \underline{68.0}\\\cline{1-1}\cline{3-12}
        \cellcolor{lightgray}\textbf{IFA \unboldmath$_{\textbf{\textit{T}}=3}$}& \cellcolor{lightgray}~& \cellcolor{lightgray}\textbf{50.6}& \cellcolor{lightgray}\textbf{58.8}& \cellcolor{lightgray}\textbf{66.3}& \cellcolor{lightgray}\textbf{69.8}& \cellcolor{lightgray}\textbf{74.0}& \cellcolor{lightgray}\textbf{74.6}& \cellcolor{lightgray}\textbf{80.1}& \cellcolor{lightgray}\textbf{82.4}& \cellcolor{lightgray}\textbf{67.8}& \cellcolor{lightgray}\textbf{71.4}\\
        \bottomrule[1pt]
    \end{tabular}
    \caption{Quantitative CD-FSS results under the mIoU (\%) evaluation metric. The best and second best results are highlighted with \textbf{bold} and \underline{underline}, respectively. $*$ means the results are reproduced by ourselves.}
    \label{tab:cd-fss}
\end{table*}

\begin{figure}[ht]
  \centering
    \includegraphics[width=1.0\linewidth]{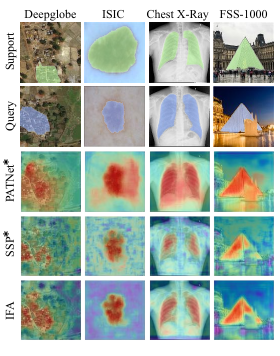}
    \vspace{-10mm}
    \caption{Qualitative results of the samples in four target datasets. From left to right, each column shows examples from Deepglobe, ISIC, Chest X-Ray, and FSS-1000. From up to down, each row shows the examples of support images with ground-truth masks (green), query images with ground-truth masks (blue), PATNet results, Our baseline (SSP) results, and Our IFA results. $*$ represents the model reproduced by ourselves. Best viewed in color.}
    \label{fig:visualization}
    \vspace{-5mm}
\end{figure}

\subsection{Comparison with State-of-the-art Methods}
In Tab.~\ref{tab:cd-fss}, we report comparisons of our proposed Iterative Few-shot Adaptor (IFA) with state-of-the-art methods. SSP~\cite{fan2022self} is initially proposed for FSS, we reproduce it for CD-FSS. The mIoU (\%) is used as the evaluation metric~\cite{min2021hypercorrelation}. We construct our ResNet~\cite{he2016deep} backbone following most compared methods. Our method surpasses state-of-the-arts consistently by large margins. Particularly on the ISIC dataset, our IFA achieves 17.7\% (1-shot) and 4.4\% (5-shot) of mIoU improvements over the previous best method.

Deepglobe presents unique challenges as every pixel in the image belongs to a well-defined category. Nevertheless, only some specific pixels are region-of-interest in the CD-FSS task, while others are treated as background. The superior performance (+9.3\% on 1-shot) of the Deepglobe dataset demonstrates that our IFA excels at complex scene datasets. ISIC and Chest X-Ray are two medical screening datasets, where the background is clean and the target region occupies a large portion of the entire image. The outstanding performances affirm that our IFA effectively establishes the category correspondence of medical domains. The difficulty of FSS-1000 arises from the significant variations between 1000 categories. Despite previous methods performing well due to the relatively small domain gap between the source domain (Pascal VOC) and FSS-1000, our IFA still outperforms priors.

It is worth noting that our IFA achieves better results on 5-shot setups compared with 1-shot, indicating that our iterative design leverages more information when provided with more support images. To better analyze and understand the superiority of IFA, we visualize the segmentation results, as shown in Fig.~\ref{fig:visualization}. Our results (5$^{th}$ row) are significantly better compared to the previous best methods (3$^{rd}$ and 4$^{th}$ row). We also visualize more
qualitative results of IFA in supplementary materials.

\subsection{Ablation Studies}
\noindent\textbf{Component-wise ablation.} We conduct ablation studies of our designed components, as shown in Tab.~\ref{tab:ab-study}. The baseline is our reproduced SSP~\cite{fan2022self}. To verify our assumption that the \textit{Support-to-Query-to-Support (S2Q2S)} stream mines more support-query correspondence, we first only introduce Bi-directional Few-shot Prediction (BFP) on the source domain and apply the fine-tuning without query supervision on target domains. The performance improvements on Deepglobe(+2.8\%) and ISIC(+13.7\%) verify the effectiveness of our bi-directional design. After deriving query from annotated support in target domains and repeating BFP three times to formulate our IFA design, we achieve further performance gains of 6.5\% and 4.0\% on two datasets, respectively. Qualitative results of SSP baseline (4$^{th}$ row) and IFA (5$^{th}$ row) are shown in Fig.~\ref{fig:visualization}.

\begin{table}[ht]
    \renewcommand\arraystretch{1.15}
    \centering
    \vspace{4mm}
    \begin{tabular}{c|c|c|c}
        \toprule[1pt]
        \multicolumn{4}{c}{Pascal VOC 2012 $\to$ Below}\\\hline
        Train on source& Adapt to target& Deepglobe& ISIC\\\hline
        Baseline& None& 41.3& 48.6\\
        Baseline$+$BFP& w/o query label& 44.1& 62.3\\
        \textbf{Baseline$+$BFP}& \textbf{IFA} \unboldmath$_{\textbf{\textit{T}}=3}$& \textbf{50.6}& \textbf{66.3}\\
        \bottomrule[1pt]
    \end{tabular}
    \vspace{1mm}
    \caption{Ablation studies for components of our method.}
    \vspace{1mm}
    \label{tab:ab-study}
\end{table}

\noindent\textbf{Fine-tuning strategy ablation.} We also compare our proposed IFA strategy with other fine-tuning strategies used in relevant tasks, as shown in Tab.~\ref{tab:ft_compare}. All the strategies are performed on the same model (trained via Baseline$+$BFP) to ensure fairness. Task-Adaptive Fine-Tuning (TAFT)~\cite{lei2022cross} is encapsulated in the meta-testing stage, which is different from our independent fine-tuning design. Specifically, TAFT introduces Kullback-Leibler divergence loss to mitigate the distance between the category prototype of the segmented query image and that of the support set. Copy-Paste Fine-Tuning (CPFT)~\cite{gao2022acrofod} is employed in few-shot object detection task. Concretely, CPFT mixes the image with the foreground region of the target domain and the background region from a random image in the source domain as the support set. It compels to adapt knowledge learned from source to target domains, which is inspired from \cite{ghiasi2021simple}. From the experimental results, we find that our IFA is the most effective fine-tuning strategy for CD-FSS.

\begin{table}[ht]
    \renewcommand\arraystretch{1.15}
    \centering
    \begin{tabular}{l|c|c}
        \toprule[1pt]
        \multicolumn{3}{c}{Pascal VOC 2012 $\to$ Below}\\\hline
        Fine-tuning strategy& Deepglobe& ISIC\\\hline      TAFT~\cite{lei2022cross}& 43.2& 64.3\\
        CPFT~\cite{gao2022acrofod}& 44.6& 54.5\\
        \textbf{IFA \unboldmath$_{\textbf{\textit{T}}=3}$}& \textbf{50.6}& \textbf{66.3}\\
        \bottomrule[1pt]
    \end{tabular}
    \caption{Ablation study on different fine-tuning strategies. We conduct the same training process on the source domain, and adopt different fine-tuning strategies on the target domains.}
    \label{tab:ft_compare}
    \vspace{-3mm}
\end{table}

\subsection{Analysis}
\noindent\textbf{Effectiveness of iteration design.} To demonstrate the effectiveness of iterating BFP multiple times, we compare the results of different times of iteration, as shown in Tab.~\ref{tab:iter}. We observe that iterating BFP leads to improved performance, and performing 3 times strikes the best balance between performance and computational cost. We assume that iterating too many times may introduce superfluous information for optimization. Therefore, performing more than 3 iterations yields performance saturation.

\begin{table}[ht]
    \renewcommand\arraystretch{1.3}
    \centering
    \resizebox{\linewidth}{!}{
    \begin{tabular}{l|ccccccccc}
        \toprule[1pt]
        \multicolumn{10}{c}{Pascal VOC 2012 $\to$ Deepglobe}\\\hline
        $\textbf{\textit{T}}$& 1& 2& 3& 4& 5& 6& 7& \textbf{8}& 9\\\hline
        mIoU& 46.6& 47.2& 50.6& 50.6& 50.8& 50.7& 50.8& \textbf{50.9}& 50.8\\
        \bottomrule[1pt]
    \end{tabular}
    }
    \caption{Quantitative comparison results of different iteration times of BFP. Some visualized results and corresponding analysis are shown in supplementary materials.}
    \label{tab:iter}
    \vspace{-1mm}
\end{table}

\noindent\textbf{Comparison with Segment Anything Model (SAM).}
We compare the results of our
IFA with SAM~\cite{kirillov2023segment} in two medical image segmentation datasets (\ie, ISIC and Chest X-Ray), as shown in Tab.~\ref{tab:sam}. The results demonstrate substantial improvements from our proposed
method over SAM,
despite our method relying on a common meta-learning paradigm. It is worth noting that our model employed in this experiment has much fewer parameters and training data than those of the SAM.

\begin{table}[ht]
    \renewcommand\arraystretch{1.25}
    \centering
    \resizebox{\linewidth}{!}{
    \begin{tabular}{l|c|c|c}
        \toprule[1pt]
        Methods& Backbone& ISIC& Chest X-Ray\\\hline
        SAM~\cite{kirillov2023segment} \textit{(zero-shot)}& ViT& 36.1& 27.8\\
       \textbf{IFA \unboldmath$_{\textbf{\textit{T}}=3}$} \textit{(1-shot)}& \textbf{Res-50}& \textbf{66.3}& \textbf{74.0}\\
        \bottomrule[1pt]
    \end{tabular}
    }
    \caption{Comparison with Segment Anything Model (SAM)~\cite{kirillov2023segment} in two medical image segmentation tasks. The results of IFA are under Res-50 backbone with 1-shot setup. The results of SAM are borrowed from \cite{chen2023dense}.}
    \label{tab:sam}
    \vspace{-1mm}
\end{table}

\noindent\textbf{Results under Domain-Shift Few-Shot Segmentation (DS-FSS).} To demonstrate the robustness of our method across different datasets, we also evaluate IFA on Domain-Shift Few-Shot Segmentation (DS-FSS) tasks following the recent work of \cite{boudiaf2021few}. Referring to the definition of CD-FSS, DS-FSS can be regarded as a special condition of CD-FSS, where the source domain is COCO-20i~\cite{nguyen2019feature} while the target domain is Pacal-5i~\cite{shaban2017one} without overlapped categories. DS-FSS is slightly simpler than four tasks in \cite{lei2022cross}, because the categories in the source domain and target domain are both daily objects. Experiment results shown in Tab.~\ref{tab:ds-fss} also surpass the previous methods by significant margins (+5.4\%).
\begin{table}
    \renewcommand\arraystretch{1.1}
    \centering
    \begin{tabular}{l|c|c|c}
        \toprule[1pt]
        \multicolumn{4}{c}{Source Domain: COCO-20i $\to$ Target Domain: Pascal-5i}\\\hline
        \multirow{2}*{Methods}&  \multirow{2}*{Backbone}& {1-shot}& {5-shot}\\\cline{3-4}
        ~& ~& \textbf{Mean}& \textbf{Mean}\\\hline
        RPMMs~\cite{yang2020prototype}& \multirow{8}*{Res-50}& 49.6& 53.8\\
        PFENet~\cite{tian2020prior}& ~& 60.8& 61.9\\
        RePRI~\cite{boudiaf2021few}& ~& 57.0& 67.9\\
        ASGNet~\cite{li2021adaptive}& ~& 57.4& 66.6\\
        HSNet~\cite{min2021hypercorrelation}& ~& 61.5& 68.4\\
        CWT~\cite{lu2021simpler}& ~& 59.4& 66.5\\
        Meta-Memory~\cite{wang2022remember}& ~& \underline{65.6}& \underline{70.1}\\
        \cellcolor{lightgray}\textbf{IFA \unboldmath$_{\textbf{\textit{T}}=3}$}& \cellcolor{lightgray}~& \cellcolor{lightgray}\textbf{71.0}& \cellcolor{lightgray}\textbf{80.9}\\\hline
        SCL~\cite{zhang2021self}& \multirow{4}*{Res-101}& 59.4& 60.3\\
        HSNet~\cite{min2021hypercorrelation}& ~& 63.2& 70.0\\
        Meta-Memory~\cite{wang2022remember}& ~& \underline{66.6}& \underline{72.7}\\
        \cellcolor{lightgray}\textbf{IFA \unboldmath$_{\textbf{\textit{T}}=3}$}& \cellcolor{lightgray}~& \cellcolor{lightgray}\textbf{79.6}& \cellcolor{lightgray}\textbf{83.4}\\
        \bottomrule[1pt]
    \end{tabular}
    \caption{Quantitative Domain-Shift Few-Shot Segmentation results under the mIoU (\%) evaluation metric. The best and second best results are highlighted with \textbf{bold} and \underline{underline}, respectively. The concrete performances of each fold are shown in supplementary materials.}
    \label{tab:ds-fss}
    \vspace{-3mm}
\end{table}

\section{Conclusion}

This paper presents an in-depth exploration of Cross-Domain Few-Shot Segmentation (CD-FSS) tasks, highlighting the critical role of fine-tuning in transferring FSS capabilities across various domains. We also uncover the challenge of overfitting due to limited data availability. With these insights, we propose a novel cross-domain fine-tuning strategy. We first design Bi-directional Few-shot Prediction (BFP) that establishes extensive category correspondence to mitigate the overfitting risk. Then we extend BFP to Iterative Few-shot Adaptor (IFA), which recursively mines support-query correspondence by maximally exploiting supervisory signals from limited data. Extensive experiments show that our designs tackle both cross-domain and overfitting challenges simultaneously and outperform state-of-the-arts by large margins. We hope that our work inspires further research in developing more effective algorithms and exploring additional facets of few-shot tasks.

\section*{Acknowledgement}
This study is supported by the Interdisciplinary Graduate Programme, Nanyang Technological University, and the Ministry of Education Singapore, under the Tier-1 scheme with project number RG18/22.
{
    \small
    \bibliographystyle{ieeenat_fullname}
    \bibliography{main}
}

\clearpage
\setcounter{page}{1}
\maketitlesupplementary

\setcounter{table}{0}
\setcounter{figure}{0}
\setcounter{section}{0}
\setcounter{equation}{0}
\renewcommand{\thetable}{A\arabic{table}}
\renewcommand{\thefigure}{A\arabic{figure}}
\renewcommand{\thesection}{A\arabic{section}}
\renewcommand{\theequation}{A\arabic{equation}}

\section{Cross-Domain Few-Shot Segmentation}

\begin{figure}[ht]
    \centering
    \includegraphics[width=1.0\linewidth]{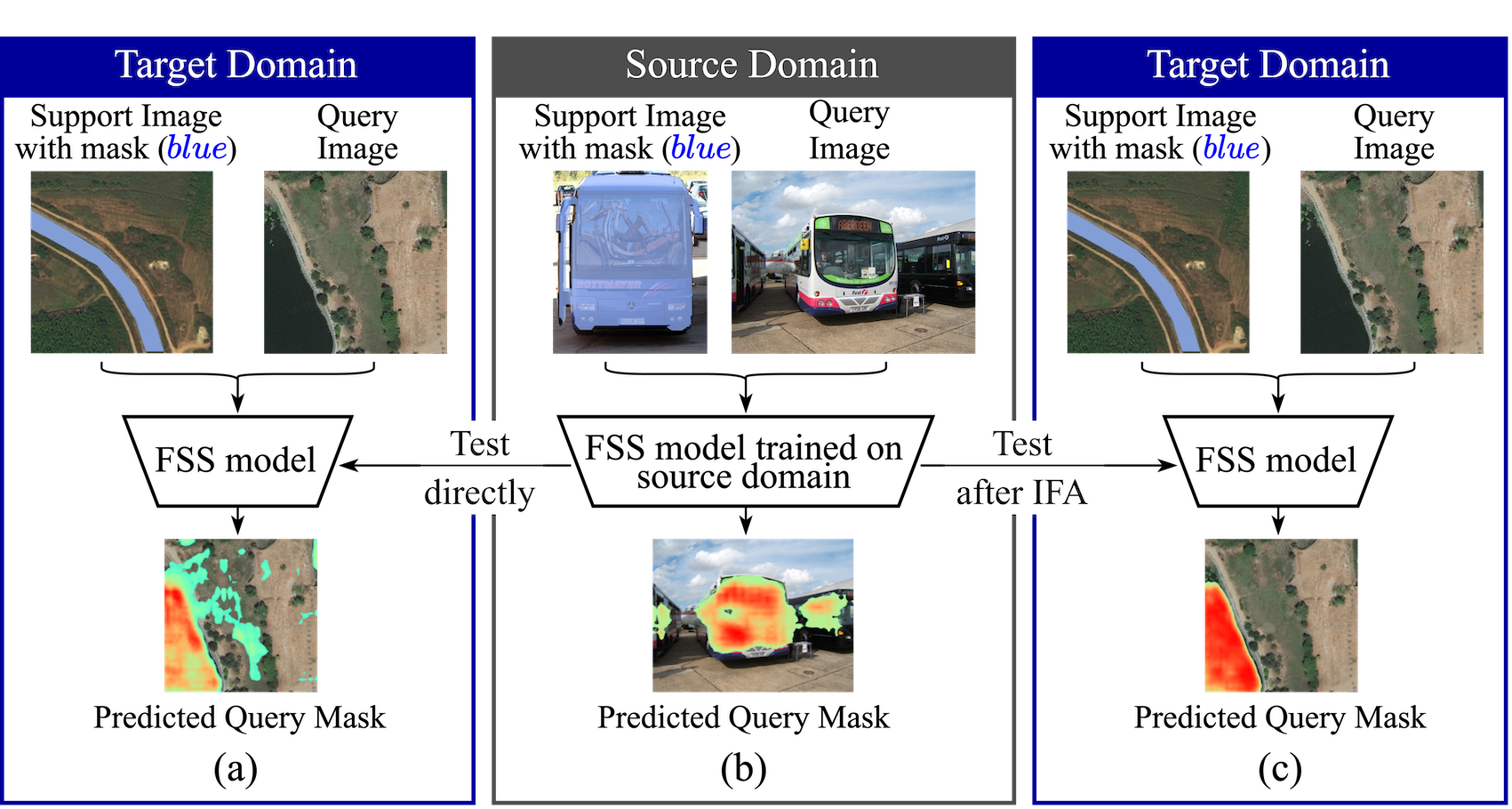}
    \caption{The formulation of Cross-Domain Few-Shot Segmentation task. (a) The segmentation performance clearly suffers from a severe drop when directly applying the trained model to target images. (b) Meta-training model on the source domain. (c) The segmentation performance is clearly improved benefiting from our proposed Iterative Few-shot Adaptor (IFA).}
    \label{fig:formulation}
\end{figure}

\noindent\textbf{Motivation.} Few-Shot Segmentation (FSS) methods rely on abundant base categories data to learn the capability of segmenting novel categories with a few exemplars~\cite{snell2017prototypical,dong2018few,tian2020prior,fan2022self,min2021hypercorrelation,zhang2021few,peng2023hierarchical}. However, collecting sufficient annotated data is infeasible in low-resource domains (\eg, satellite images and medical screenings), thus the FSS pipeline is no longer suitable. Cross-Domain Few-Shot Segmentation (CD-FSS)~\cite{lei2022cross} proposes a possible solution for the above challenge. Specifically, it aims to meta-train models on a source domain (\eg, Pascal VOC~\cite{everingham2010pascal}) with abundant accessible data, and adapt trained models to low-resource domains with a small support set (refer to Fig.~\ref{fig:formulation}). CD-FSS prposes an efficient solution for some specific downstream tasks (\eg TB detection and wildlife conservation), where collecting substantial data is laborious, costly, and may raise privacy issues~\cite{lei2022cross}. The CD-FSS pipeline eases the efforts of collecting and annotating large amounts of data in low-resource target domains.

\noindent\textbf{Comparison with Domain-Shift Few-Shot Segmentation.} Although Boudiaf~\etal~\cite{boudiaf2021few} propose a Domain-Shift Few-Shot Segmentation (DS-FSS) setting and claim it is more realistic compared with FSS benchmarks by using data in a different domain for evaluation. Nevertheless, both base and novel categories are from daily object datasets. Therefore, the images in DS-FSS setting are easy to collect in large quantities, which is not a challenge in real application scenarios. Moreover, only utilizing daily object categories for evaluation can not exhibit the generalization capacity entirely. Consequently, Lei~\etal~\cite{lei2022cross} explore specific domains within CD-FSS task, which is more challenging: \textit{(i)} satellite images and medical screenings are difficult to collect due to expensive cost and privacy agreements respectively, \textit{(ii)} tiny or scarce objects are always neglected even in few-shot datasets~\cite{shaban2017one, nguyen2019feature}.

\noindent\textbf{Existing problem.} A feasible solution for CD-FSS is to perform meta-training on an annotation-abundant domain (\eg, Pascal VOC~\cite{everingham2010pascal}) and subsequently transfer to target data-limited ones. However, the learned models often suffer from a severe performance drop when directly applied to a different domain as shown in Fig.~\ref{fig:formulation}(a), and such a problem cannot be easily tackled by simply improving the few-shot capability. Because the core challenge is brought by the clear domain gap, we propose Iterative Few-shot Adaptor (IFA). In such a way, a model mines more support-query correspondence with extremely limited data, and generalized the capability of segmenting novel categories to target domains (refer to Fig.~\ref{fig:formulation}(c)).

\section{More Details of Applying SSP}
\noindent We use Self-Support Prototype (SSP)~\cite{fan2022self} as our baseline, and also inherit its specific designs. To mitigate the effect form cluttered background, we incorporate the adaptive self-support background prototype~\cite{fan2022self} into our framework. Besides, we also use self-support refinement to achieve more accurate predictions, which is effective in \cite{fan2022self}. For fair comparisons, we conduct all the experiments of the main paper with the same setup.

\section{Experimental Result}

\noindent\textbf{Results under Domain-Shift Few-Shot Segmentation.} The concrete performances under Domain-Shift Few-Shot Segmentation (DS-FSS)
are shown in Tab.~\ref{tab:ds-fss} We observe that our IFA surpasses the previous best method with large margins in all folds, which validates the robustness and effectiveness of our designs.

\begin{figure*}[h]
    \centering
    \includegraphics[width=1.0\linewidth]{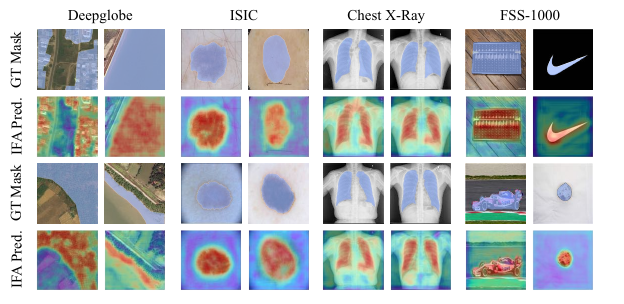}
    \caption{More qualitative results of Iterative Few-shot Adaptor (IFA) in four target datasets. Best viewed in color.}
    \label{fig:more_vis}
\end{figure*}

\noindent\textbf{More visualization results.} We display more qualitative prediction results of our proposed Iterative Few-shot Adaptor (IFA), as shown in Fig.~\ref{fig:more_vis}. It is obvious that IFA successfully transfers the capability learned from the source domain to four target domains. In Deepglobe~\cite{demir2018deepglobe}, our IFA segments different categories from satellite images with similar accuracy. Besides, IFA segments medical screenings accurately from ISIC~\cite{codella2019skin} and Chest X-Ray~\cite{tschandl2018ham10000}. For FSS-1000~\cite{li2020fss}, IFA predicts target objects across different types (\eg, logos, foods, and objects) with satisfying results.

\section{Analysis}

\begin{figure}[ht]
    \centering
    \vspace{-3mm}
    \includegraphics[width=0.75\linewidth]{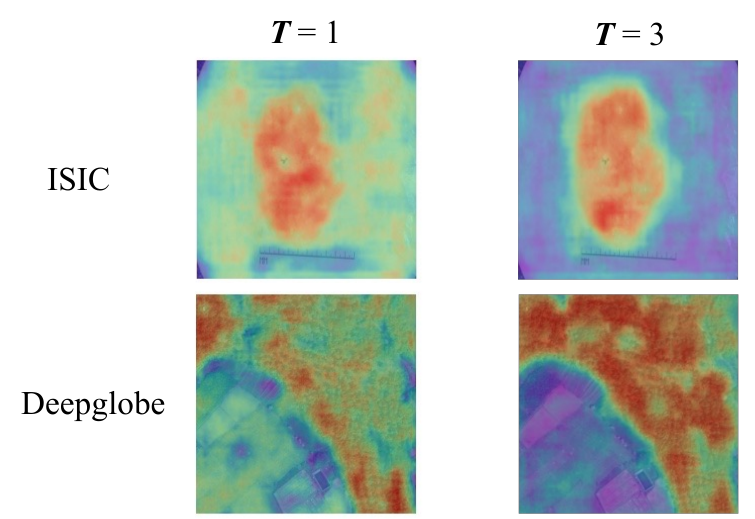}
    \caption{Visualized segmentation result (examples from ISIC and Deepglobe) after $\textbf{\textit{T}}$ times of recursive prediction.}
    \label{fig:iter_result}
\end{figure}

\noindent\textbf{Effectiveness of iteration design.} We also visualize the prediction after different iterations of BFP, as shown in Fig.~\ref{fig:iter_result}. We observe that the iterative design increases the prediction results significantly. This validates our assumption outlined in the main paper that our IFA provides extensive information for mining support-query correspondence.

\noindent\textbf{Iterative computing cost.} The iteration design is only used in fine-tuning, which only increases a little bit costs. Specifically, the time of each fine-tuning epoch increases to 12.87s from 4.82s, and the GPU memory occupies extra 29.3
Mb. The training and testing stages are unaffected.

\begin{table}[H]
    \renewcommand\arraystretch{1.1}
    \centering
    \tabcolsep=0.3em
    \resizebox{\linewidth}{!}{
    \begin{tabular}{l|ccccccccc}
        \toprule[1pt]
        \multicolumn{10}{c}{Pascal VOC 2012 $\to$ Deepglobe}\\\hline
        $\lambda_{s'}$& 0.1& 0.2& 0.3& \textbf{0.4}& 0.5& 0.6& 0.7& 0.8& 0.9\\\hline
        mIoU& 50.2& 50.1& 50.3& \textbf{50.6}& 50.4& 50.4& 50.3& 50.2& 49.8\\\hline\hline
        $\lambda_{i}$& \textbf{0.1}& 0.2& 0.3& 0.4& 0.5& 0.6& 0.7& 0.8& 0.9\\\hline
        mIoU& \textbf{50.6}& 50.1& 49.9& 49.9& 50.3& 50.2& 49.8& 49.7& 50.2\\\hline\hline
        $\lambda_{bs}$& 0.1& \textbf{0.2}& 0.3& 0.4& 0.5& 0.6& 0.7& 0.8& 0.9\\\hline
        mIoU& 50.4& \textbf{50.6}& 50.5& 50.1& 49.9& 50.2& 50.3& 50.2& 50.5\\\hline\hline
        $\lambda_{bq}$& & 0.5& & \textbf{1.0}& & 1.5& & 2.0& \\\hline
        mIoU& & 50.0& & \textbf{50.6}& & 49.6& & 49.7& \\\hline
    \end{tabular}
    }
    \caption{Impact of the hyper-parameters $\lambda_{s'}$, $\lambda_{i}$, $\lambda_{bs}$, and $\lambda_{bq}$ which denote the weight of $\mathcal{L}_{s'}$, $\mathcal{L}_{i}$, $\mathcal{L}_{bs}$, and $\mathcal{L}_{bq}$ in main paper.}
    \label{tab:my_label}
\end{table}

\noindent\textbf{Hyper-parameter value.} We conduct experiments to determine the value of $\lambda_{s'}$, $\lambda_{i}$, $\lambda_{bs}$, and $\lambda_{bq}$, which balance the loss terms in main paper. Specifically, we first meta-train on Pascal VOC and use IFA to transfer the learned model to Deepglobe (1-shot setup with Res-50 backbone). From Tab.~\ref{tab:my_label}, we observe that the best performance is achieved when $\lambda_{s'}=0.4$, $\lambda_{i}=0.1$, $\lambda_{bs}=0.2$, and $\lambda_{bq}=1.0$ thus we determine the value of these hyper-parameters in our all experiments.

\noindent\textbf{Validating Gestalt principle on four target datasets.} Fan~\etal~\cite{fan2022self} prove the Gestalt principle~\cite {koffka2013principles} existing in the daily object dataset by the cosine similarity statistics. Following their methods, we also conduct experiments in four target domains of the CD-FSS task. Statistic information is shown in Tab.~\ref{tab:ssp-principle}. We can find that pixels belonging to the same object have much higher similarity than the cross-object pixels in all datasets, thus the Gestalt principle and SSP method are still effective. For the FSS-1000 dataset, we use all the images to compute the cosine similarity. For the remaining datasets, we randomly pick 200 images in each category for computation, except for picking 100 images of class 2 in ISIC due to data insufficiency.

\begin{table}[ht]
    \renewcommand\arraystretch{1.1}
    \centering
    \tabcolsep=0.3em
    \resizebox{\linewidth}{!}{
    \begin{tabular}{cc|cc|cc|cc}
        \toprule[1pt]
        \multicolumn{8}{c}{ForeGround Pixels Similarity}\\\hline
        \multicolumn{2}{c|}{Deepglobe}& \multicolumn{2}{c|}{ISIC}& \multicolumn{2}{c|}{Chest X-Ray}& \multicolumn{2}{c}{FSS-1000}\\\hline
        cross& intra& cross& intra& cross& intra& cross& intra\\
        0.497& \textbf{0.552}&0.512& \textbf{0.526}& 0.528& \textbf{0.554}& 0.494& \textbf{0.563} \\
        \bottomrule[1pt]
    \end{tabular}
    }
    \caption{Cosine similarity for cross/intra-object pixels in four CD-FSS datasets.}
    \label{tab:ssp-principle}
\end{table}

\begin{table*}[h]
    \renewcommand\arraystretch{1.15}
    \centering
    \begin{tabular}{l|c|cccc|c|cccc|c}
        \toprule[1pt]
        \multicolumn{12}{c}{Source Domain: COCO-20i $\to$ Target Domain: Pascal-5i}\\\hline
        \multirow{2}*{Methods}&  \multirow{2}*{Backbone}& \multicolumn{5}{c|}{1-shot}& \multicolumn{5}{c}{5-shot}\\\cline{3-12}
        ~& ~& fold-0& fold-1& fold-2& fold-3& \textbf{Mean}& fold-0& fold-1& fold-2& fold-3& \textbf{Mean}\\\hline
        RPMMs~\cite{yang2020prototype}& \multirow{8}*{Res-50}& 36.3& 55.0& 52.5& 54.6& 49.6& 40.2& 58.0& 55.2& 61.8& 53.8\\
        PFENet~\cite{tian2020prior}& ~& -& -& -& -& 60.8& -& -& -& -& 61.9\\
        RePRI~\cite{boudiaf2021few}& ~& 52.4& \underline{64.3}& 65.3& 71.5& 63.3& 57.0& 68.0& 70.4& 76.2& 67.9\\
        ASGNet~\cite{li2021adaptive}& ~& 42.5& 58.7& 65.5& 63.0& 57.4& 53.7& \underline{69.8}& 67.1& 75.9& 66.6\\
        HSNet~\cite{min2021hypercorrelation}& ~& 48.7& 61.5& 63.0& 72.8& 61.5& 58.2& 65.9& \underline{71.8}& \underline{77.9}& 68.4\\
        CWT~\cite{lu2021simpler}& ~& 53.5& 59.2& 60.2& 64.9& 59.4& 60.3& 65.8& 67.1& 72.8& 66.5\\
        Meta-Memory~\cite{wang2022remember}& ~& \underline{57.4}& 62.2& \underline{68.0}& \underline{74.8}& \underline{65.6}& \underline{65.7}& 69.2& 70.8& 75.0& \underline{70.1}\\
        \cellcolor{lightgray}\textbf{Ours\unboldmath$_{T=3}$}& \cellcolor{lightgray}~& \cellcolor{lightgray}\textbf{61.9}& \cellcolor{lightgray}\textbf{71.4}& \cellcolor{lightgray}\textbf{68.7}& \cellcolor{lightgray}\textbf{82.0}& \cellcolor{lightgray}\textbf{71.0}& \cellcolor{lightgray}\textbf{73.2}& \cellcolor{lightgray}\textbf{82.1}& \cellcolor{lightgray}\textbf{80.4}& \cellcolor{lightgray}\textbf{88.0}& \cellcolor{lightgray}\textbf{80.9}\\\hline
        SCL~\cite{zhang2021self}& \multirow{4}*{Res-101}& 43.1& 60.3& 66.1& 68.1& 59.4& 43.3& 61.2& 66.5& 70.4& 60.3\\
        HSNet~\cite{min2021hypercorrelation}& ~& 46.3& \underline{64.7}& 67.7& \underline{74.2}& 63.2& 59.1& 69.0& \underline{73.4}& 78.7& 70.0\\
        Meta-Memory~\cite{wang2022remember}& ~& \underline{59.4}& 64.3& \underline{70.8}& 72.0& \underline{66.6}& \underline{67.2}& \underline{72.7}& 72.0& \underline{78.9}& \underline{72.7}\\
        \cellcolor{lightgray}\textbf{Ours\unboldmath$_{T=3}$}& \cellcolor{lightgray}~& \cellcolor{lightgray}\textbf{71.3}& \cellcolor{lightgray}\textbf{77.1}& \cellcolor{lightgray}\textbf{80.0}& \cellcolor{lightgray}\textbf{89.8}& \cellcolor{lightgray}\textbf{79.6}& \cellcolor{lightgray}\textbf{77.7}& \cellcolor{lightgray}\textbf{84.6}& \cellcolor{lightgray}\textbf{80.3}& \cellcolor{lightgray}\textbf{90.8}& \cellcolor{lightgray}\textbf{83.4}\\
        \bottomrule[1pt]
    \end{tabular}
    \caption{More detailed Quantitative comparison results on Domain-Shift Few-Shot Segmentation problem using mIoU (\%) evaluation metric. The best and second best results are highlighted with \textbf{bold} and \underline{underline}, respectively.}
    \label{tab:ds-fss}
\end{table*}


\end{document}